\documentclass[10pt,twocolumn,letterpaper]{article}

\usepackage{cvpr}              

\usepackage[dvipsnames]{xcolor}

\definecolor{cvprblue}{rgb}{0.21,0.49,0.74}
\usepackage[pagebackref,breaklinks,colorlinks,citecolor=cvprblue]{hyperref}
\usepackage{url}            
\usepackage{booktabs}       
\usepackage{amsfonts}       
\usepackage{nicefrac}       
\usepackage{microtype}      
\usepackage{xcolor}         
\usepackage{multirow}
\usepackage{graphicx}
\usepackage{amsmath}
\usepackage{amssymb}
\usepackage{threeparttable}

\def\paperID{15237} 
\def\confName{CVPR}
\def\confYear{2024}

\title{CompactFlowNet: Efficient Real-time Optical Flow Estimation on Mobile Devices}

\author{
Andrei Znobishchev\\
Picsart AI Research (PAIR)\\
{\tt\small anznobishchev@gmail.com}
\and
Valerii Filev\\
Picsart AI Research (PAIR)\\
{\tt\small valerii.filev@picsart.com}\\
\and
Oleg Kudashev\\
Picsart AI Research (PAIR)\\
{\tt\small oleg.kudashev@picsart.com}\\
\and
Nikita Orlov\\
Picsart AI Research (PAIR)\\
{\tt\small nikita.orlov@picsart.com}\\
\and
Humphrey Shi\\
Picsart AI Research (PAIR)\\
SHILabs @ U of Oregon $\&$ UIUC\\
{\tt\small shihonghui3@gmail.com}\\
}

\begin{document}
\maketitle
\begin{abstract}
We present CompactFlowNet, the first real-time mobile neural network for optical flow prediction, which involves determining the displacement of each pixel in an initial frame relative to the corresponding pixel in a subsequent frame. Optical flow serves as a fundamental building block for various video-related tasks, such as video restoration, motion estimation, video stabilization, object tracking, action recognition, and video generation. While current state-of-the-art methods prioritize accuracy, they often overlook constraints regarding speed and memory usage. Existing light models typically focus on reducing size but still exhibit high latency, compromise significantly on quality, or are optimized for high-performance GPUs, resulting in sub-optimal performance on mobile devices. This study aims to develop a mobile-optimized optical flow model by proposing a novel mobile device-compatible architecture, as well as enhancements to the training pipeline, which optimize the model for reduced weight, low memory utilization, and increased speed while maintaining minimal error. Our approach demonstrates superior or comparable performance to the state-of-the-art lightweight models on the challenging KITTI and Sintel benchmarks. Furthermore, it attains a significantly accelerated inference speed, thereby yielding real-time operational efficiency on the iPhone 8, while surpassing real-time performance levels on more advanced mobile devices.
\end{abstract}    
\section{Introduction}
Within the scope of computer vision tasks, optical flow estimation functions as a fundamental component integral to numerous algorithms that range from action recognition \citep{simonyan2014recognition}, \citep{yang2021unsupervised}, target tracking \citep{shuang2020fine}, and video stabilization \citep{yu2020learning}, \citep{10089609}, to more composite tasks like autonomous navigation \citep{chen2019learning}, \citep{wang2018correlation} and human-robot interaction \citep{chang2019improved}. The objective of optical flow is to determine pixel-wise mapping between a source image and a target image, expressed as a 2D displacement field. 

Early research works have proven the effectiveness of approaches based on energy minimization, as highlighted in multiple studies \citep{horn1981determining}, \citep{brox2010large}, \citep{zach2007duality}, \citep{kroeger2016fast}. Later, a series of convolutional neural networks (CNNs) \citep{dosovitskiy2015flownet}, \citep{ilg2017flownet} were suggested for learning optical flow. Those works proved that deep learning methods can reach similar performance to traditional methods in terms of accuracy, while being significantly faster.

Recent optical flow models \citep{xu2017accurate}, \citep{sun2018pwc}, \citep{hur2019iterative}, \citep{yang2019volumetric}, \citep{teed2020raft}, \citep{huang2022flowformer}, \citep{shi2023flowformer++} have capitalized on the concept of feature warping and cost volume computation to improve performance. Feature warping allows these models to handle large displacements by warping the features of the second image towards the first using an initial flow estimate. In addition, the use of cost volumes, which are 4D tensors storing the matching costs of features at different potential displacements, provides a more detailed and granular matching strategy. This has enabled these models to make more accurate and precise flow predictions, improving the overall optical flow estimation. Despite the notable advancements in the accuracy of optical flow prediction with recurrent \citep{teed2020raft}, \citep{chen2023mfcflow} and transformer-based \citep{jaegle2021perceiver}, \citep{jiang2021learning}, \citep{xu2022gmflow}, \citep{zhao2022global}, \citep{huang2022flowformer}, \citep{shi2023flowformer++} models in recent years, a critical examination of state-of-the-art methodologies reveals a frequent disregard for certain constraints, including model size, latency, and memory utilization. While a series of attempts have been initiated to enhance the compactness of optical flow models \citep{ranjan2017optical}, \citep{hui18liteflownet}, \citep{hui2020liteflownet3}, \citep{kong2020fdflownet}, \citep{kong2021fastflownet}, these efforts often demonstrate a compromise on accuracy, a lack of consideration for inference speed, or a limitation to testing on high-performance hardware. 

Modern applications, however, necessitate real-time inference capabilities on mobile devices, as it provides numerous advantages, with the most noticeable one being the enhancement of overall inference latency. This is achieved via eliminating the necessity to upload data to a server and wait for inference results, resulting in the app being able to respond to user requests more promptly. Other benefits include the elimination of the need for server maintenance and the capability to function with limited or no internet connectivity. Furthermore, keeping user data on the device helps mitigate privacy issues. 

With the advent of powerful mobile processors, efficient memory management, and energy-optimized designs, mobile devices have become capable of handling computationally demanding tasks \citep{lee2019device}. Deploying deep learning models on mobile devices entails meeting specific requirements to ensure optimal performance. Firstly, the number of parameters in the models should be carefully considered. Mobile devices often have resource limitations, making it essential to reduce parameter count. Studies indicate that models with parameter count limited to a few million parameters are suitable for mobile deployment \citep{howard2017mobilenets}, \citep{ignatov2018ai}. Secondly, the memory footprint of the models is a critical concern, as limited memory on mobile devices can lead to performance issues. To ensure efficient memory utilization, models with memory footprints ranging from tens to hundreds of megabytes are preferable for mobile deployment \citep{ignatov2018ai}. Furthermore, latency, or inference time, is a vital consideration for mobile applications. Since optical flow is typically applied to video-related tasks \citep{fan2018end}, \citep{niklaus2018context}, \citep{chang2019improved}, \citep{yu2020learning}, \citep{chan2021basicvsr}, \citep{10089609}, real-time inference speed is often necessary.

Hence, the purpose of this research is to take into account the aforementioned requirements and propose a neural network architecture for optical flow prediction suitable for efficient real-time inference on mobile devices. In summary, our contributions are the following:
\begin{itemize}
    \item A novel architecture is suggested that prioritizes model compactness and accelerates inference speed on mobile devices, thereby addressing the crucial challenge of efficient deployment.
    \item A heavier backbone architecture recently designed specifically for mobile applications is utilized. It improves accuracy without compromising the inference time on mobile devices.
    \item A meticulous evaluation of the speed and memory efficiency of lightweight optical flow methods is undertaken across various contemporary mobile devices and diverse input resolutions for the first time.
\end{itemize}

To our knowledge, this is the first compact and memory-efficient model for optical flow prediction with real-time inference performance on mobile devices.

\section{Related work}
\label{gen_inst}

Various approaches have been suggested for optical flow estimation. Most of them consider a two-frame case when a model aims to predict optical flow given two consecutive frames. Although methods that incorporate a larger number of frames for prediction exist \citep{janai2018unsupervised}, \citep{8658399}, \citep{godet2021starflow}, \citep{chen2023mfcflow}, \citep{shi2023videoflow} they impose a heavier demand on computational time and memory resources. Thus, only two-frame methods are investigated further.

\paragraph{Early work on optical flow}
Optical flow has been well studied with classical optimization methods, and can be formulated as an energy minimization problem. This method, first introduced in \citep{horn1981determining}, focuses on the optimization of an energy function within a coarse-to-fine framework. However, the inherent computational heaviness of such energy optimization techniques poses significant challenges for their implementation in real-time applications even when high-performance hardware is available.

Utilizing CNNs has instigated a paradigm shift in the field of computer vision. \citep{dosovitskiy2015flownet} introduced two CNN models, FlowNetS and FlowNetC, showing the capability of these networks to directly estimate optical flow from raw image data. While these models did not surpass the accuracy of state-of-the-art methods, they demonstrated significant improvement of inference speed. Subsequent iterations, specifically FlowNet2 \citep{ilg2017flownet}, further enhanced the accuracy of optical flow estimation, nearing the performance metrics of energy minimization methods, but with significantly superior computational efficiency. FlowNet2, in particular, is a cascade of FlowNet variants, each of which incrementally refines the optical flow field. However, this complex network configuration was not optimized for its size and computational efficiency, resulting in 160M parameters and 640 megabytes memory footprint \citep{kong2021fastflownet}. 

Further generation of CNN-based models, i.e. PWC-Net \citep{sun2018pwc} and IRR-PWC \citep{hur2019iterative}, further refined this approach by introducing pyramid processing and iterative refinement respectively, leading to higher precision and more detailed flow maps. These methods employ an iterative coarse-to-fine approach which entails estimating optical flow on low resolution, subsequently leveraging this initial estimation to inform predictions at the higher resolution levels. Feature correspondence is identified with a correlation module within a predetermined window at each level, a strategy that proves to be significantly more efficient than establishing correspondence amongst all pixels at the original resolution. Even though the search range has a fixed size, large displacement flow can still be estimated due to feature warping and the coarse-to-fine approach. Such methods deliver a decent compromise among accuracy, memory-efficiency and computational speed, indicating their potential applicability in practical scenarios.

\paragraph{Recent state-of-the-art methods}

Cutting-edge methodologies for optical flow prediction extend beyond the conventional CNNs and exploit the latest developments in computer vision. RAFT \citep{teed2020raft}, for instance, constructs multi-scale 4D correlation volumes for all pairs of pixels and employs a recurrent unit for iterative updates of flow estimates. While this strategy enhances the accuracy of the model, it also imposes significant computational burden. For instance, inference on images of resolution 1024x448 requires 107 milliseconds and more than 2.5 gigabytes of GPU memory \citep{sun2022disentangling}. Alternatively, other methodologies have harnessed the potential of attention mechanism and transformer-based strategies \citep{jaegle2021perceiver}, \citep{jiang2021learning}, \citep{xu2022gmflow}, \citep{huang2022flowformer}, \citep{shi2023flowformer++} and \citep{luo2023gaflow}. GMFlowNet \citep{zhao2022global}, for instance, employs a global matching strategy in conjunction with patch-based overlapping attention, thereby achieving superior quality compared to RAFT. However, this enhancement is accompanied by an increase in latency by 30$\%$. 

Despite their impressive accuracy, state-of-the-art models are incapable of real-time inference, even on powerful GPUs. Furthermore, their suitability for mobile applications is compromised due to their extensive parameter size and peak memory usage during inference. This is largely attributed to the construction of substantial cost volumes and the utilization of attention mechanisms. Additionally, the inherent nature of attention mechanisms results in a quadratic surge in GPU memory consumption proportional to the increase in the area of input frames.

\paragraph{Fast and lightweight models}
There is a high demand for fast optical flow prediction in practical applications. Typically, optical flow is used for video processing, which involves estimating flow multiple times per second. Notably, while optical flow constitutes a crucial component of a plethora of video-related algorithms, it merely represents one of the blocks of these methodologies, with the remaining parts often demanding significant inference time. This additionally highlights importance of fast inference speed of optical flow. 

As one of the pioneering compact models in the field, SpyNet \citep{ranjan2017optical} employs a spatial pyramid network architecture for optical flow prediction, which effectively approximates large displacements. Due to a small size of merely 1.2M parameters, SpyNet was often used for video-related applications such as video enhancement \citep{xue2019video}, frame interpolation \citep{niklaus2018context} and video super-resolution \citep{chan2021basicvsr}. However, it is worth noting that despite its compactness, SpyNet was not specifically designed for swift inference. Consequently, its latency is 47.4 milliseconds, which is slower than the 32.2 milliseconds (NVIDIA 1080Ti GPU, 1024x436 resolution) latency of more modern PWC-Net \citep{kong2020fdflownet}. Another model named LiteFlowNet \citep{hui18liteflownet} offers several key improvements over SpyNet. This network introduces a lightweight, yet more accurate and efficient architecture that is specifically designed for speed and outperforms SpyNet in terms of both quality and inference time. However, when compared to PWC-Net, LiteFlowNet is slower with a latency of 53.2 miliseconds (NVIDIA 1080Ti GPU, 1024x436 resolution) \citep{kong2020fdflownet}. PWC-Net-small \citep{sun2018pwc} proposed by the authors of PWC-Net is a light version of the original model. Its reduced size is achieved via dropping DenseNet connections, which also makes it more suitable for memory limited applications. FDFlowNet \citep{kong2020fdflownet} employs a U-shape network for multi-scale feature extraction and a partial fully connected structure with dilated convolution for flow estimation. This approach is about 2 times faster than the original PWC-Net. Finally, FastFlowNet \citep{kong2021fastflownet} introduces several architectural improvements over PWC-Net. A key feature is the Head Enhanced Pooling Pyramid (HEPP), providing more efficient feature extraction. Also, it incorporates the center dense dilated correlation (CDDC), a modification to the standard cost volume that reduces calculation while keeping large search radius. Lastly, it introduces a Shuffle Block Decoder, a more compact and efficient decoding mechanism than that used by PWC-Net. These specific design choices accelerate the inference process while maintaining high-quality predictions.

\section{Method}
\label{method}

As mentioned previously, CNN-based optical flow models and PWC-Net in particular provide a desirable balance between accuracy, memory efficiency and inference speed. Therefore, the architecture of PWC-Net was adopted as a prototype for the current study. Next, a profiling analysis was conducted to identify the computationally demanding components of PWC-Net's architecture. These identified blocks were then optimized to accelerate the inference process, reduce memory usage, and decrease the model's overall size. The profiling method is explained in Section \ref{profiling_method}, while architecture modifications are covered in Section \ref{arch_redesign}. The adoption of a lighter model unavoidably leads to a reduction in its quality. To address this issue and restore the desired level of accuracy, a series of steps were undertaken. Firstly, since the introduction of PWC-Net, numerous studies have focused on developing mobile feature extractors. Section \ref{backbone_replacement} explores alternative options for replacing the original backbone with a more efficient module. Moreover, a distillation pipeline is introduced to narrow the accuracy gap between the lighter model and the heavier model. See details in Section \ref{training_mod}.

\subsection{Profiling method}
\label{profiling_method}

Initially, all the models were developed using Pytorch \citep{NEURIPS2019_bdbca288} and trained on NVIDIA GPU. Nevertheless, given that our research concentrates on examining on-mobile inference, it was essential to deploy the models on mobile devices. One of the frameworks utilized for this purpose is TensorFlow Lite (TFLite) \citep{tflite2019}. The choice of TFLite was influenced by its capability of GPU inference and its compatibility with both iOS and Android, thereby ensuring adaptability. There is no direct method to convert models from PyTorch to TFLite. Instead, this process necessitates several stages, which includes transforming the model from PyTorch to ONNX, subsequently converting from ONNX to TensorFlow \citep{tensorflow2016}, and finally from TensorFlow to TFLite.

The profiling process was conducted on the iPhone 8 to ensure that real-time inference capabilities can be achieved even for a relatively old smartphone model. The resolution of input images was set to 512x512, deemed as an appropriate resolution setting for practical applications. Benchmarking results for more modern devices and other input resolutions can be found in Section \ref{main_results}.

\subsection{Architecture redesign}
\label{arch_redesign}
\begin{figure}
  \centering
  \includegraphics[scale=0.045]{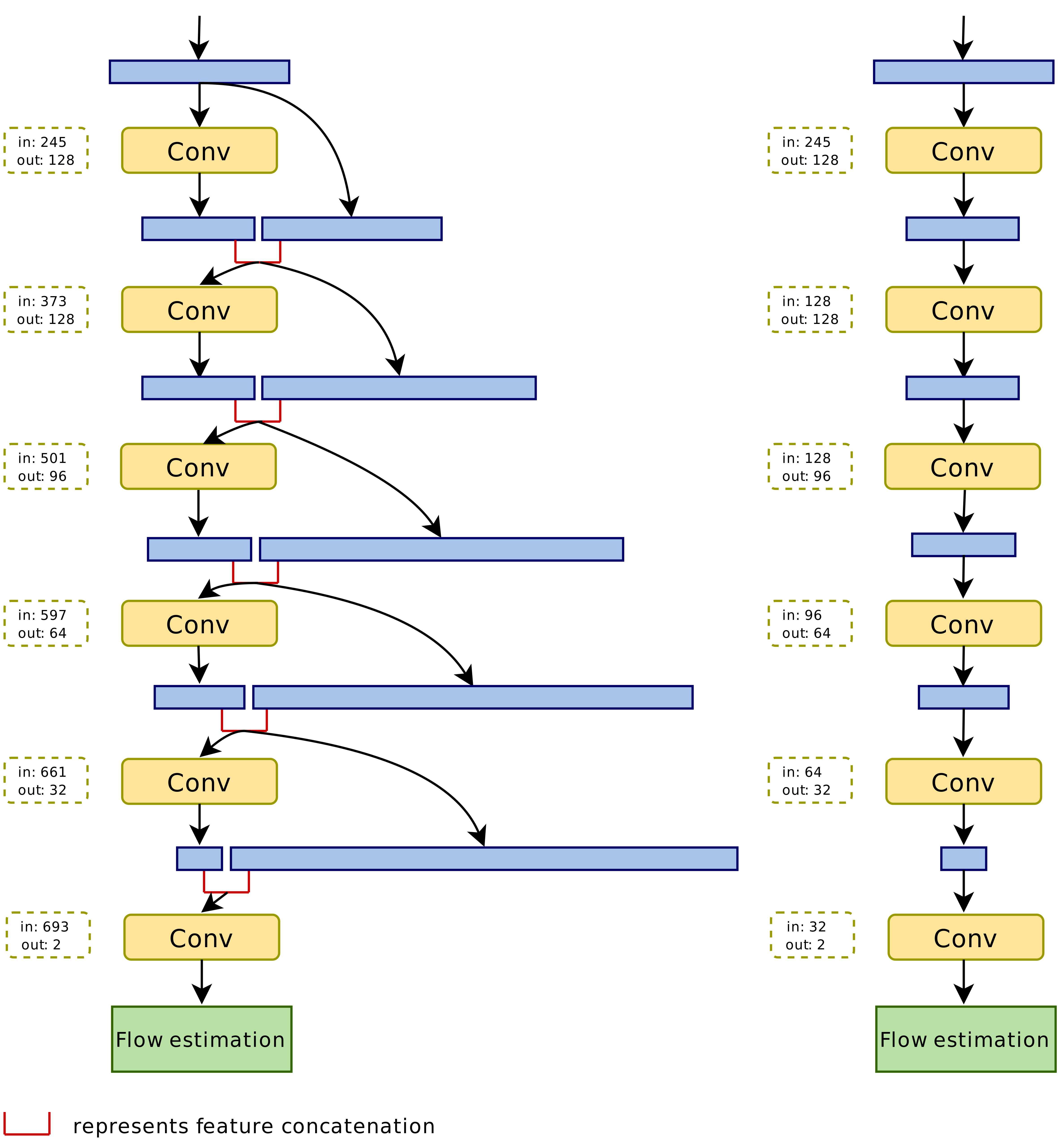}
  \caption{Comparison of flow estimator structures: densely connected (left) and sequentially connected (right). The indicated number of input and output channels corresponds to the flow estimator block operating at the lowest resolution.}
  \label{fig:figure_flow_estimator_blocks}
\end{figure}

PWC-Net encompasses three main blocks: the feature extractor, flow estimator, and flow refiner. An investigation into the structure of PWC-Net reveals that the DenseNet approach, i.e. the feature concatenation employed in the flow estimator block, significantly impacts the inference performance of the network. The flow estimator block plays a dominant role in terms of parameter count, FLOPs, and latency, as indicated in Table~\ref{table_pwcnet_profiling}. Furthermore, the reliance on feature concatenation in this block necessitates the retention of large feature tensors in memory, as depicted in Figure~\ref{fig:figure_flow_estimator_blocks}. Additionally, the feature concatenation within the flow estimator block has implications for the flow refiner, as it relies on the final feature map generated by the flow estimator. Hence, this introduces an increase in parameter count and latency within the flow refiner as well. Therefore, the elimination of feature concatenation in the flow estimator block becomes critical for developing a mobile-compatible architecture. Consequently, a sequentially connected structure was adopted for the flow estimator to address this need, see Figure~\ref{fig:figure_flow_estimator_blocks}.

\begin{table}[ht]
\caption{Computational profiling of PWC-Net conducted on iPhone 8 using input images with a resolution of 512x512}
\label{table_pwcnet_profiling}
\centering
\small
\begin{tabular}{l l l l} 
\toprule
 & Feature extractor & Flow estimator & Flow refiner \\
\midrule
Params (M)                 & 1.67 & 6.05 & 1.03 \\
FLOPs (G)                           & 3.0 & 26.8 & 18.5 \\
Latency, ms                         & 19 & 131 & 62 \\
\bottomrule
\end{tabular}
\end{table}

Furthermore, a reduction in the depth of the convolutional pyramid within the flow refiner was implemented, utilizing only four layers instead of the original seven. The number of channels for these layers was set to 128, 64, 32, and 2. 

Additionally, taking inspiration from the approach proposed in \citep{howard2017mobilenets}, depthwise separable convolutions were adopted in both the flow estimator and flow refiner blocks. By decomposing the standard convolution into separate depth-wise and point-wise operations, depth-wise separable convolutions offer a notable reduction in computational requirements and parameter count. This enhanced efficiency makes them particularly well-suited for resource-constrained environments while maintaining a high level of accuracy.

\subsection{Backbone replacement}
\label{backbone_replacement}
The MobileNet models \citep{howard2017mobilenets}, \citep{sandler2018mobilenetv2}, \citep{howard2019searching} have revolutionized mobile and embedded vision applications by introducing depthwise separable convolutions, inverted residuals, linear bottlenecks, and network architecture search with the new operations, thereby significantly enhancing the performance and computational efficiency of convolutional neural networks on resource-constrained devices. 
EfficientNet models \citep{howard2017mobilenets}, \citep{sandler2018mobilenetv2}, \citep{tan2019efficientnet}, \citep{tan2021efficientnetv2} have presented a novel compound scaling method that uniformly scales the depth, width, and resolution of convolutional neural networks, optimizing them for both accuracy and efficiency. Building upon the breakthroughs of MobileNet and EfficientNet, a plethora of new models have emerged \citep{tan2019mixconv}, \citep{wu2019fbnet}, \citep{tan2019mnasnet}, \citep{lee2019energy}, \citep{lee2020centermask}, leveraging their innovative components, compound scaling, and neural architecture search. 

Taking into account mobile-compatibility constraints described earlier, it is essential that the new backbone architecture maintains less than 4.5 million parameters. This ensures that the complete optical flow model does not exceed 5 million parameters, yielding a model size under 10 megabytes when converted to float16. A comparative study was conducted among several state-of-the-art architectures that meet these criteria, including  ReXNet \citep{han2021rethinking}, MobileNetV3 \citep{howard2019searching}, HardCoReNas \citep{nayman2021hardcore}, MobileViT \citep{mehta2021mobilevit}, MobileViTv3 \citep{wadekar2022mobilevitv3} and EfficientFormer \citep{li2022efficientformer}. Despite the compact nature of MobileViT and MobileViTv3, their performance does not meet the requirements of the current work as the backbone applied to two input images requires more than 40 milliseconds on the iPhone 8 at the desired resolution of 512x512, thus impeding the full optical flow model from achieving the real-time 25 FPS inference. This problem also extends to the other attention-based model, EfficientFormer. Although this method demonstrates impressive efficiency on ImageNet resolution images \citep{deng2009imagenet},  it struggles to satisfy the established constraints when applied to larger input sizes, due to its quadratic scalability with respect to the area of the input. Thus, only CNN-based models were tested, see Table \ref{table_backbones} and Table \ref{table_backbones2}. For this comparison, only the backbone was substituted in the original PWC-Net model. Then, each new model was trained on a mixture of datasets, see Section \ref{exps} for details.

\begin{table}[ht]
\caption{Parameter count and inference speed across backbone architectures. Model latency corresponds to the inference speed of the full optical flow model. Latency is calculated as an average over 100 runs performed after 100 warm-up runs on iPhone 8 using input images with a
resolution of 512x512}
\label{table_backbones}
\centering
\small
\begin{tabular}{l l l} 
\toprule
Backbone & Params (M) & Model latency, ms  \\
\midrule
Original (PWC-Net)                       & $\textbf{1.67}$ & $\textbf{39}$  \\
HardCoRe-Nas-B                 & $\underline{2.67}$  & 41 \\
MobileNetV3 (large)             & 3.12 & $\underline{40}$ \\
HardCoRe-Nas-C                 & 3.09 & 43 \\
ReXNet(×1.0)                    & 3.54 & 45 \\
\bottomrule
\end{tabular}
\end{table}

\begin{table}[ht]
\caption{Accuracy metrics across backbone architectures. The figures enclosed in parentheses represent the results of networks that were trained on the same dataset. AEPE metric is used}
\label{table_backbones2}
\centering
\small
\begin{tabular}{l l l l} 
\toprule
Backbone & \multicolumn{2}{c}{Sintel Final} & KITTI 2015 \\
\cmidrule(lr){2-3}
 & train & validation & train  \\
\midrule
Original (PWC-Net)                       & (2.15) & 3.78 & (2.23) \\
HardCoRe-Nas-B                 & (1.71) & 3.08 & (1.91) \\
MobileNetV3 (large)             & (1.66) & $\underline{2.87}$ & (1.83) \\
HardCoRe-Nas-C                & (1.67) & 2.96 & (1.82) \\
ReXNet(×1.0)                   & (1.65) & $\textbf{2.80}$ & (1.80) \\
\bottomrule
\end{tabular}
\end{table}

Interestingly, these models exhibit similar performance metrics when evaluated on the Sintel and KITTI datasets. Consequently, among the proposed alternatives, MobileNetV3 stands out by offering the fastest inference speed and demonstrating a significant enhancement in quality relative to the original architecture.

\subsection{Distillation}
\label{training_mod}
The original PWC-Net, modified with MobileNetV3 as the backbone, was employed as a heavy teacher model. CompactFlowNet, the streamlined model as described in Section \ref{arch_redesign}, was trained as a student model within a distillation pipeline. The loss function used for training is the following: 
\begin{equation}
{\mathcal{L}} = {\mathcal{L}}_{sup} + \gamma * {\mathcal{L}}_{dist}
\end{equation}
where ${\mathcal{L}}_{sup}$ is the supervision loss calculated as per-pixel L2 norm between the predicted flow and the ground-truth flow on each resolution level, ${\mathcal{L}}_{dist}$ is the distillation loss reflecting the discrepancy between the predicted flow of the teacher and student models, and $\gamma$ is the weight of the distillation loss set to 0.1.

\section{Experiments}
\label{exps}

\subsection{Implementation details}
\label{impl_details}
FlyingChairs \citep{dosovitskiy2015flownet} and FlyingThings3D \citep{mayer2016large} datasets have been typically used for optical flow model pre-training. Recently, a novel dataset Autoflow \citep{sun2021autoflow} has been rendered providing more diverse shapes, motions and appearances of moving objects leading to better accuracy on pre-training, which was validated for PWC-Net and RAFT. Besides, a more diverse and complex augmentation scheme has been suggested. Accordingly, we have integrated a broader range of augmentations with more adjustable parameters. The augmentation pipeline incorporates color jitter, Gaussian noise, eraser transform, random crop and pad, horizontal flip, vertical flip and random scale. See specific augmentation parameters in Supplementary Materials.

Moreover, authors of \citep{sun2022disentangling} advocate that models such as Flow1D \citep{xu2021high}, SeparableFlow \citep{zhang2021separable}, PWC-Net, IRR-PWC and RAFT, which employ a OneCycle learning rate and gradient clipping, have shown performance improvements with these training techniques. The OneCycle learning rate schedule starts with a low learning rate, incrementally escalates to reach a peak value, and subsequently regresses back to the value below the initial learning rate. Utilization of gradient clipping improves stability of training and has been reported to deliver better performance for PWC-Net and IRR-PWC \citep{hofinger2020improving}. Thus, the reported training modification were adopted in this study as well.

The original PWC-Net with MobileNetV3 as the backbone is first pre-trained on Autoflow using the OneCycle learning rate for 1.2 million iterations. Next, the heavy model is fine-tuned on a mixture of FlyingThings3D \citep{mayer2016large}, Sintel \citep{butler2012mpi}, KITTI 2015 \citep{Menze2015CVPR}, HD1K \citep{kondermann2016hci} and VIPER \citep{Richter_2017} datasets with a piece-wise learning rate, as described in \citep{sun2019models} for 750000 iterations.  Subsequently, CompactFlowNet is initialized with backbone weights taken from the trained heavy model. Finally, CompactFlowNet is trained in a distillation pipeline with the OneCycle learning rate for 1.2 million iterations on a mixture of Autoflow, FlyingThings3D, Sintel, KITTI 2015, HD1K and VIPER datasets. The resultant model does not exhibit overfitting to a specific dataset; instead, it demonstrates generalization capabilities, enabling it to estimate optical flow across diverse domains. Recent state-of-the-art models, such as RAFT and Flowformer++, were considered as potential options for teacher models as well. However, their integration noticeably hampers the training process without yielding discernible performance enhancements for the student model.

Weight decay is set to 0 for pre-training and distillation, and to 1e-5 for fine-tuning.

\subsection{Main results}
\label{main_results}
Evaluation results of CompactFlowNet, and its comparison to the reference PWC-Net model and to the lightweight methods are summarized in Table~\ref{table_main_results_sintel} and  Table~\ref{table_main_results_kitty}. Furthermore, running times can be found in Table~\ref{table_main_results_speed}. Finally, the peak memory usage is shown in Table~\ref{table_memory_footprint}. Performance is measured directly on device for all of the listed methods, i.e. all of the models were implemented in PyTorch and converted to TFLite for on-device inference. Latency is calculated as an average over 100 runs performed after 100 warm-up runs. Inference capabilities of the models under investigation are demonstrated using input images of 512x512, 436x1024 (the resolution of Sintel images), and 1080x1920 (Full HD) resolutions. This strategy effectively represents a range of computational demands, thereby providing a comprehensive analysis of model performance under varying operational conditions. PWC-Net+ corresponds to the regular PWC-Net model trained within the enhanced pipeline of \cite{sun2022disentangling}. 

\begin{figure}[!htb]
  \centering
  \includegraphics[scale=0.12]{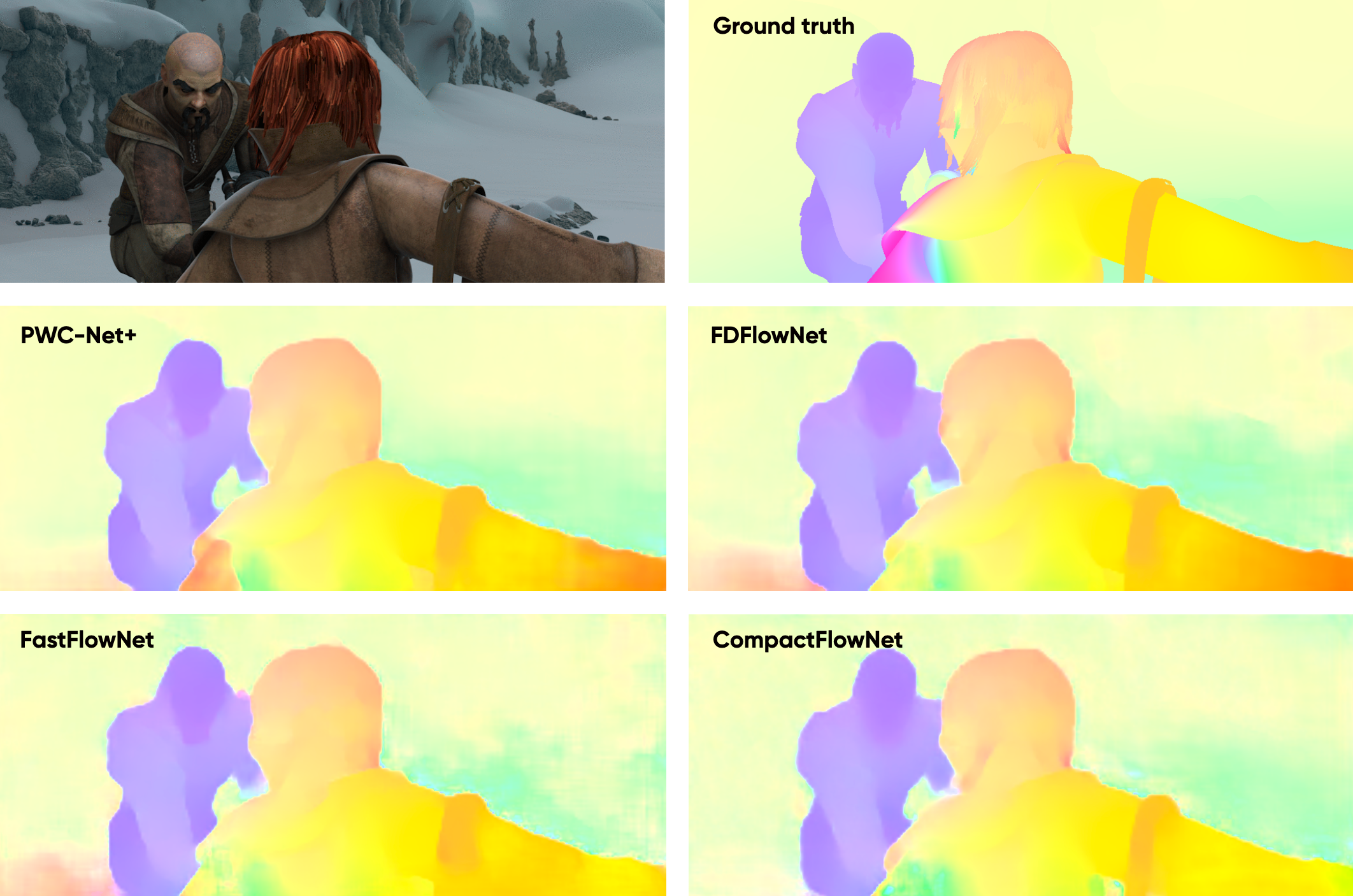}
  \caption{Visualized results on Sintel Clean test set.}
  \label{fig:figure_results_sintel_clean}
\end{figure}

\begin{figure}
  \centering
  \includegraphics[scale=0.12]{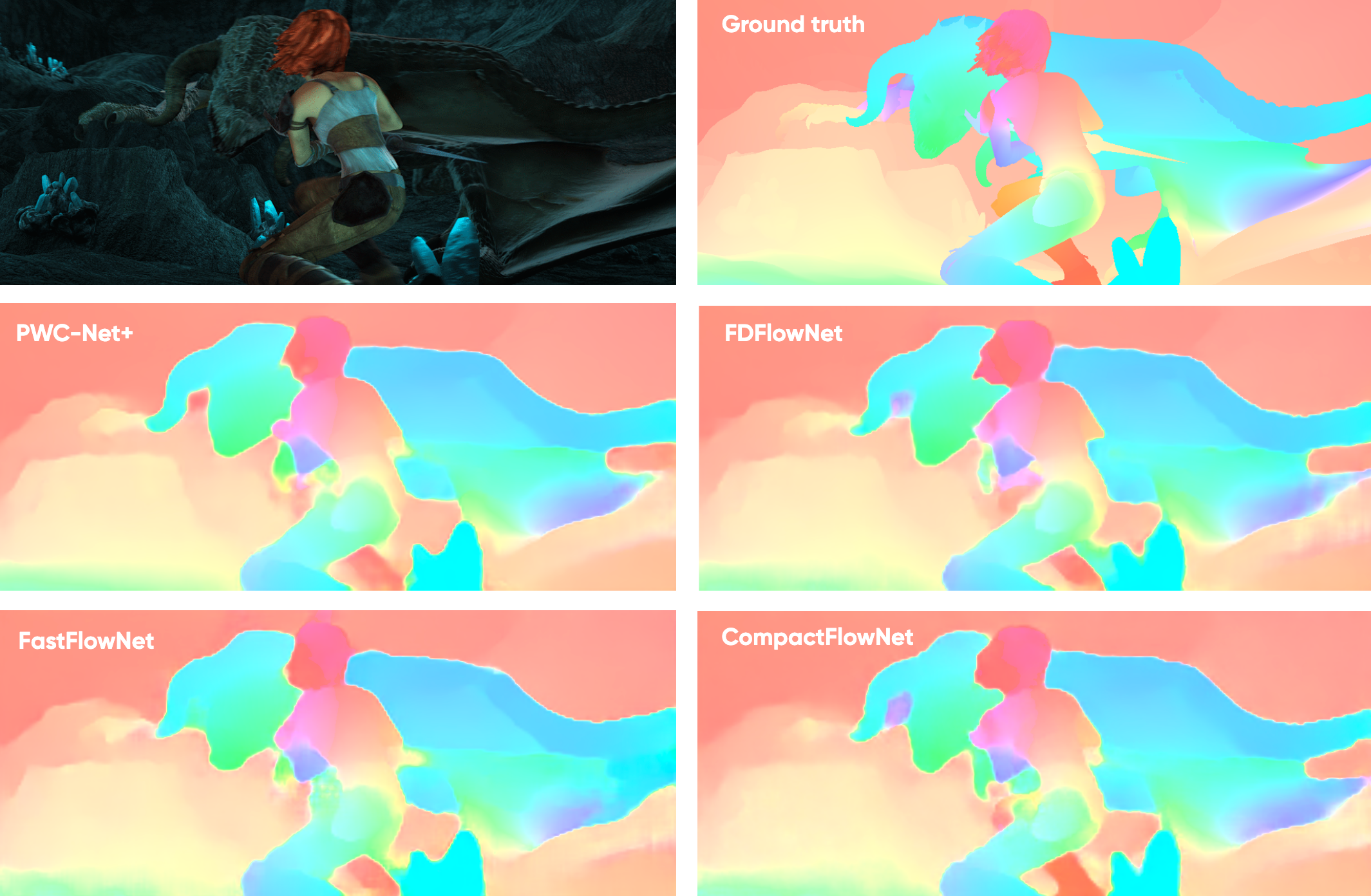}
  \caption{Visualized results on Sintel Final test set.}
  \label{fig:figure_results_sintel_final}
\end{figure}

\begin{figure}
  \centering
  \includegraphics[scale=0.22]{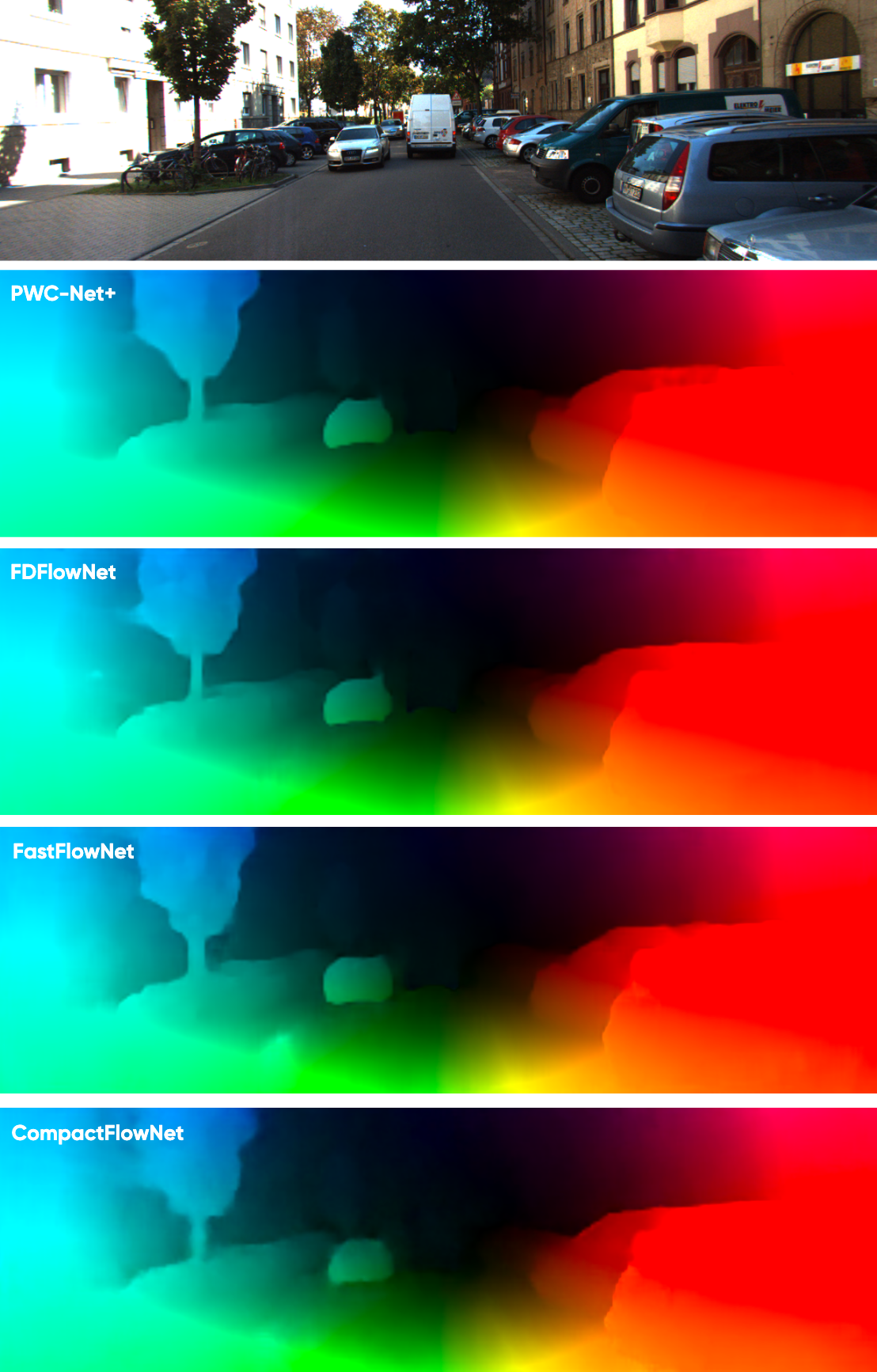}
  \caption{Visualized results on KITTI 2015 test set.}
  \label{fig:figure_results_kitti}
\end{figure}

In addition, see Figure~\ref{fig:figure_results_sintel_clean}, Figure~\ref{fig:figure_results_sintel_final} and Figure~\ref{fig:figure_results_kitti} to visually review performance of different lightweight models.

It is important to note that the comparison excludes computationally intensive state-of-the-art methods such as RAFT \citep{teed2020raft}, Flowformer++ \citep{shi2023flowformer++}, Videoflow \citep{shi2023videoflow} and others as they are currently impractical for deployment in the mobile devices domain. Some of their respective papers contain assessments of the speed and memory utilization on high-performance GPUs. Besides, the paper \citep{sun2022disentangling} provides an in-depth examination of latency and peak-memory utilization concerning PWC-Net and RAFT.

\begin{table}[ht]
\caption{Performance comparison of lightweight optical flow models on the Sintel Clean and Sintel Final datasets. The figures enclosed in parentheses represent the results of networks that were trained on the same dataset. The end-point error is the default accuracy metric}
\label{table_main_results_sintel}
\centering
\small
\begin{tabular}{l l l l l l l l} 
\toprule
{Method} & \multicolumn{2}{c}{Sintel Clean} & \multicolumn{2}{c}{Sintel Final}   \\
\cmidrule(lr){2-3}
\cmidrule(lr){4-5}

                          & train & test & train & test    \\
\midrule
PWC-Net-small                  & (2.83) & - & (4.08) & -  \\
SPyNet                  & (3.17) & 6.64 & (4.32) & 8.36  \\
LiteFlowNet                  & (1.64) & 4.86 & (2.23) & 6.09  \\
FDFlowNet                  & (1.8) & $\textbf{3.71}$ & (1.93) & $\textbf{5.11}$  \\
FastFlowNet                  & (2.08) & 4.89 & (2.71) & 6.08  \\
CompactFlowNet                  & (1.55) & $\underline{4.43}$ & (1.94) & $\underline{5.55}$  \\
\bottomrule
\end{tabular}
\end{table}

\begin{table}[ht]
\caption{Performance comparison of lightweight optical flow models on the KITTI dataset. The figures enclosed in parentheses represent the results of networks that were trained on the same dataset. The end-point error is the default accuracy metric}
\label{table_main_results_kitty}
\centering
\small
\begin{tabular}{l l l l l l l} 
\toprule
{Method} & \multicolumn{3}{c}{KITTI 2015}  \\
\cmidrule(lr){2-4}
                        & train & train (f1-all) & test (f1-all)   \\
\midrule
PWC-Net-small                  & - & - & - & \\
SPyNet                  & (4.13) & - & 35.07  \\
LiteFlowNet                  & (2.16) & (8.16) & $\underline{10.24}$ \\
FDFlowNet                  & (1.56) & (6.36) & $\textbf{9.38}$ \\
FastFlowNet                  & (2.13) & (8.21) & 11.22  \\
CompactFlowNet                  & (1.91) & (7.50) & 14.64 \\
\bottomrule
\end{tabular}
\end{table}

\begin{table*}[t]
\caption{Running time (ms) across lightweight optical flow models. Latency is measured as an average over 100 runs performed after 100 warm-up runs for the following devices: iPhone 8  \(|\) iPhone XR  \(|\) iPhone 12  \(|\) iPhone 14 Pro}
\label{table_main_results_speed}
\centering
\small
\begin{tabular}{l p{4cm} p{4cm} p{4cm}} 
\toprule
{Method} & \multicolumn{3}{c}{Input resolution}  \\
\cmidrule(lr){2-4}
 & \multicolumn{1}{c}{512x512} & \multicolumn{1}{c}{436x1024} & \multicolumn{1}{c}{1080x1920} \\
\midrule  
PWC-Net+                  & \multicolumn{1}{c}{212 \(|\) 	115 \(|\) 	114 \(|\) 	62}	& \multicolumn{1}{c}{343 \(|\) 	198 \(|\) 	154 \(|\) 	102} & 	\multicolumn{1}{c}{1787 \(|\) 	1440 \(|\) 	875 \(|\) 	474} \\
PWC-Net-small             & \multicolumn{1}{c}{96 \(|\) 	58 \(|\) 	45 \(|\) 	\textbf{30}} &	\multicolumn{1}{c}{167 \(|\) 	103 \(|\) 	72 \(|\) 	51} 	& \multicolumn{1}{c}{926 \(|\) 	553 \(|\) 	406 \(|\) 	235} \\
FDFlowNet                 & \multicolumn{1}{c}{127 \(|\) 	72 \(|\) 	53 \(|\) 	\textbf{37}}	& \multicolumn{1}{c}{218 \(|\) 	125 \(|\) 	93 \(|\) 	63} &	\multicolumn{1}{c}{1187 \(|\) 	658 \(|\) 	504 \(|\) 	292} \\
FastFlowNet               & \multicolumn{1}{c}{60 \(|\) 	\textbf{36} \(|\) 	\textbf{32} \(|\) 	\textbf{16}} &	\multicolumn{1}{c}{91 \(|\) 	54 \(|\) 	\textbf{41} \(|\) 	\textbf{28} } &	\multicolumn{1}{c}{433 \(|\) 	248 \(|\) 	214 \(|\) 	137} \\
CompactFlowNet            & \multicolumn{1}{c}{\textbf{40} \(|\) 	\textbf{26} \(|\) 	\textbf{19} \(|\) 	\textbf{13}} &	\multicolumn{1}{c}{71 \(|\) 	\textbf{41} \(|\) 	\textbf{34} \(|\) 	\textbf{23} } &	\multicolumn{1}{c}{349 \(|\) 	209 \(|\) 	179 \(|\) 	114} \\
\bottomrule
\end{tabular}
\begin{tablenotes}
      \small
      \centering \item[a] Running times indicated in bold font correspond to latency compatible with real-time inference.
\end{tablenotes}
\end{table*}

\begin{table}[!htb]
\caption{Peak memory footprint (MB)}
\label{table_memory_footprint}
\centering
\small
\begin{tabular}{l c c c} 
\toprule
\multirow{2}{*}{Method} & \multicolumn{3}{c}{Input resolution}  \\
\cmidrule(lr){2-4}
 & 512x512 & 436x1024 & 1080x1920 \\
\midrule  
PWC-Net+                  & 239 &	327 &	879 \\
PWC-Net-small             & $\underline{151}$ &	$\underline{205}$ &	$\underline{529}$ \\
FDFlowNet                 & 185 &	257 &	659 \\
FastFlowNet               & 261 &	271 &	535 \\
CompactFlowNet            & \textbf{118} &	\textbf{146} &	\textbf{370} \\
\bottomrule
\end{tabular}
\end{table}

\subsection{Discussion}
\label{discussion}

CompactFlowNet provides high quality optical flow prediction while exhibiting superior speed and the lowest memory utilization on all evaluated devices compared to other lightweight approaches. As noted in Section \ref{gen_inst}, SpyNet stands out as a widely embraced option for downstream video processing tasks. Consequently, CompactFlowNet, characterized by its superior quality and optimized inference capabilities, can function as a more adept foundational element for a spectrum of video-related algorithms. PWC-Net-small \cite{sun2018pwc}, akin to CompactFlowNet, enhances memory efficiency by eliminating DenseNet connections within the flow estimator module. However, without additional measures to reduce computational demand, its latency remains prohibitive for real-time inference. In addition, CompactFlowNet achieves superior accuracy due to its advanced backbone and the implementation of the distillation pipeline. With respect to the quality metrics CompactFlowNet aligns closely with that of FDFlowNet and PWC-Net+, despite these latter models being significantly more time-consuming on mobile devices. FastFlowNet, while presenting a satisfactory inference speed, underperforms on the Sintel test and is slower on mobile devices compared to CompactFlowNet by a margin of 20-50$\%$. Remarkably, CompactFlowNet is the pioneer in providing online inference speed on iPhone 8 and iPhone XR, handling relatively high resolutions of 512x512 and 436x1024 respectively. Despite online inference on larger resolutions posing a significant challenge, CompactFlowNet successfully achieves nearly 10 FPS on iPhone 14 Pro when processing FullHD resolution input images.

Moreover, when it comes to memory usage, CompactFlowNet surpasses other lightweight optical flow models, as illustrated in Table~\ref{table_memory_footprint}. FDFlowNet incorporates partial dense connections within its flow estimator, a feature that significantly increases peak GPU memory usage during the inference process. Similarly, the architecture of FastFlowNet is less optimal in terms of memory utilization. Although the specific memory demands are dependent on the application, CompactFlowNet's reduced memory consumption increases its adaptability and suitability for various uses. 

With its superior attributes, including less memory usage, quicker inference speed, and high quality, CompactFlowNet stands out as the optimal selection for real-world applications.

\subsection{Ablations}
\label{subsection_ablations}

A sequence of experiments are carried out to validate the significance of the proposed alterations to the training scheme. The initial step involves training CompactFlowNet on a combination of Autoflow, FlyingThings3D, Sintel, KITTI 2015, HD1K, and VIPER datasets. Then, we pre-load backbone weights obtained from the heavy counterpart model as described in Section \ref{impl_details} and train CompactFlowNet the same way. Further, we conduct an experiment with distillation, see Section \ref{training_mod}. Evidently, the proposed modifications contribute positively towards the benchmark metrics, see Table~\ref{table_main_results_ablation}. 

\begin{table}[!htb]
    \caption{Ablation study. AEPE metric is used}
    \label{table_main_results_ablation}
    \centering
    \footnotesize
    \begin{tabular}{l l l l l l} 
    \toprule
    Pre-trained & Distillation & \multicolumn{2}{c}{Sintel Final} & KITTI 2015 \\
    \cmidrule(lr){3-4}
                              backbone &  & train & validation & train  \\
    \midrule  
    - & - & (2.39) & 4.05 & (2.14) & \\
    \checkmark & - & (2.03) & 3.64 & (2.06) & \\
    \checkmark & \checkmark & (1.94) & $\textbf{3.36}$ & (1.91) & \\
    \bottomrule
    \end{tabular}
\end{table}

\section{Conclusions}
We have introduced CompactFlowNet, a new efficient and compact model for high-quality dense optical flow estimation specifically designed for use on mobile devices with real-time inference capabilities. Given that optical flow serves as a fundamental component in numerous video-based tasks, including video restoration, video inpainting, video effects application, diffusion-based video generation and others, this model enables these algorithms to be executed directly on mobile devices, considerably reducing associated costs.

{
    \small
    \bibliographystyle{ieeenat_fullname}
    \bibliography{list_of_references}

\begin{thebibliography}{72}
\providecommand{\natexlab}[1]{#1}
\providecommand{\url}[1]{\texttt{#1}}
\expandafter\ifx\csname urlstyle\endcsname\relax
  \providecommand{\doi}[1]{doi: #1}\else
  \providecommand{\doi}{doi: \begingroup \urlstyle{rm}\Url}\fi

\bibitem[tfl()]{tflite2019}
Google llc. tensorflow lite.
\newblock \url{https://www.tensorflow.org/lite}.
\newblock Accessed: 2019-04-08.

\bibitem[Abadi et~al.(2016)Abadi, Barham, Chen, Chen, Davis, Dean, Devin, Ghemawat, Irving, Isard, Kudlur, Levenberg, Monga, Moore, Murray, Steiner, Tucker, Vasudevan, Warden, Wicke, Yu, and Zheng]{tensorflow2016}
Martin Abadi, Paul Barham, Jianmin Chen, Zhifeng Chen, Andy Davis, Jeffrey Dean, Matthieu Devin, Sanjay Ghemawat, Geoffrey Irving, Michael Isard, Manjunath Kudlur, Josh Levenberg, Rajat Monga, Sherry Moore, Derek~G. Murray, Benoit Steiner, Paul Tucker, Vijay Vasudevan, Pete Warden, Martin Wicke, Yuan Yu, and Xiaoqiang Zheng.
\newblock Tensorflow: A system for large-scale machine learning.
\newblock In \emph{12th USENIX Symposium on Operating Systems Design and Implementation (OSDI 16)}, pages 265--283, 2016.

\bibitem[Ashar et~al.(2023)Ashar, Sadiq, Mohiuddin, Ashraf, Imran, and Ullah]{10089609}
Rana Ashar, Burhan Sadiq, Hira Mohiuddin, Saniya Ashraf, Muhammad Imran, and Anayat Ullah.
\newblock Video stabilization using raft-based optical flow.
\newblock In \emph{2023 International Conference on Robotics and Automation in Industry (ICRAI)}, pages 1--5, 2023.

\bibitem[Brox and Malik(2010)]{brox2010large}
Thomas Brox and Jitendra Malik.
\newblock Large displacement optical flow: descriptor matching in variational motion estimation.
\newblock \emph{IEEE transactions on pattern analysis and machine intelligence}, 33\penalty0 (3):\penalty0 500--513, 2010.

\bibitem[Butler et~al.(2012)Butler, Wulff, Stanley, and Black]{butler2012mpi}
D Butler, Jonas Wulff, G Stanley, and M Black.
\newblock Mpi-sintel optical flow benchmark: Supplemental material.
\newblock In \emph{MPI-IS-TR-006, MPI for Intelligent Systems (2012}. Citeseer, 2012.

\bibitem[Chan et~al.(2021)Chan, Wang, Yu, Dong, and Loy]{chan2021basicvsr}
Kelvin~CK Chan, Xintao Wang, Ke Yu, Chao Dong, and Chen~Change Loy.
\newblock Basicvsr: The search for essential components in video super-resolution and beyond.
\newblock In \emph{Proceedings of the IEEE/CVF Conference on Computer Vision and Pattern Recognition}, pages 4947--4956, 2021.

\bibitem[Chang et~al.(2019)Chang, Tejero-de Pablos, and Harada]{chang2019improved}
Jen-Yen Chang, Antonio Tejero-de Pablos, and Tatsuya Harada.
\newblock Improved optical flow for gesture-based human-robot interaction.
\newblock In \emph{2019 International Conference on Robotics and Automation (ICRA)}, pages 7983--7989. IEEE, 2019.

\bibitem[Chen et~al.(2019)Chen, Praveen, Priyantha, Muelling, and Dolan]{chen2019learning}
Yilun Chen, Palanisamy Praveen, Mudalige Priyantha, Katherina Muelling, and John Dolan.
\newblock Learning on-road visual control for self-driving vehicles with auxiliary tasks.
\newblock In \emph{2019 IEEE Winter Conference on Applications of Computer Vision (WACV)}, pages 331--338. IEEE, 2019.

\bibitem[Chen et~al.(2023)Chen, Zhu, Shi, Zhang, Zhang, Zhang, and Li]{chen2023mfcflow}
Yonghu Chen, Dongchen Zhu, Wenjun Shi, Guanghui Zhang, Tianyu Zhang, Xiaolin Zhang, and Jiamao Li.
\newblock Mfcflow: A motion feature compensated multi-frame recurrent network for optical flow estimation.
\newblock In \emph{Proceedings of the IEEE/CVF Winter Conference on Applications of Computer Vision}, pages 5068--5077, 2023.

\bibitem[Deng et~al.(2009)Deng, Dong, Socher, Li, Li, and Fei-Fei]{deng2009imagenet}
Jia Deng, Wei Dong, Richard Socher, Li-Jia Li, Kai Li, and Li Fei-Fei.
\newblock Imagenet: A large-scale hierarchical image database.
\newblock In \emph{2009 IEEE conference on computer vision and pattern recognition}, pages 248--255. Ieee, 2009.

\bibitem[Dosovitskiy et~al.(2015)Dosovitskiy, Fischer, Ilg, Hausser, Hazirbas, Golkov, Van Der~Smagt, Cremers, and Brox]{dosovitskiy2015flownet}
Alexey Dosovitskiy, Philipp Fischer, Eddy Ilg, Philip Hausser, Caner Hazirbas, Vladimir Golkov, Patrick Van Der~Smagt, Daniel Cremers, and Thomas Brox.
\newblock Flownet: Learning optical flow with convolutional networks.
\newblock In \emph{Proceedings of the IEEE international conference on computer vision}, pages 2758--2766, 2015.

\bibitem[Fan et~al.(2018)Fan, Huang, Gan, Ermon, Gong, and Huang]{fan2018end}
Lijie Fan, Wenbing Huang, Chuang Gan, Stefano Ermon, Boqing Gong, and Junzhou Huang.
\newblock End-to-end learning of motion representation for video understanding.
\newblock In \emph{Proceedings of the IEEE Conference on Computer Vision and Pattern Recognition}, pages 6016--6025, 2018.

\bibitem[Godet et~al.(2021)Godet, Boulch, Plyer, and Le~Besnerais]{godet2021starflow}
Pierre Godet, Alexandre Boulch, Aur{\'e}lien Plyer, and Guy Le~Besnerais.
\newblock Starflow: A spatiotemporal recurrent cell for lightweight multi-frame optical flow estimation.
\newblock In \emph{2020 25th International Conference on Pattern Recognition (ICPR)}, pages 2462--2469. IEEE, 2021.

\bibitem[Han et~al.(2021)Han, Yun, Heo, and Yoo]{han2021rethinking}
Dongyoon Han, Sangdoo Yun, Byeongho Heo, and YoungJoon Yoo.
\newblock Rethinking channel dimensions for efficient model design.
\newblock In \emph{Proceedings of the IEEE/CVF conference on Computer Vision and Pattern Recognition}, pages 732--741, 2021.

\bibitem[Hofinger et~al.(2020)Hofinger, Bul{\`o}, Porzi, Knapitsch, Pock, and Kontschieder]{hofinger2020improving}
Markus Hofinger, Samuel~Rota Bul{\`o}, Lorenzo Porzi, Arno Knapitsch, Thomas Pock, and Peter Kontschieder.
\newblock Improving optical flow on a pyramid level.
\newblock In \emph{Computer Vision--ECCV 2020: 16th European Conference, Glasgow, UK, August 23--28, 2020, Proceedings, Part XXVIII}, pages 770--786. Springer, 2020.

\bibitem[Horn and Schunck(1981)]{horn1981determining}
Berthold~KP Horn and Brian~G Schunck.
\newblock Determining optical flow.
\newblock \emph{Artificial intelligence}, 17\penalty0 (1-3):\penalty0 185--203, 1981.

\bibitem[Howard et~al.(2019)Howard, Sandler, Chu, Chen, Chen, Tan, Wang, Zhu, Pang, Vasudevan, et~al.]{howard2019searching}
Andrew Howard, Mark Sandler, Grace Chu, Liang-Chieh Chen, Bo Chen, Mingxing Tan, Weijun Wang, Yukun Zhu, Ruoming Pang, Vijay Vasudevan, et~al.
\newblock Searching for mobilenetv3.
\newblock In \emph{Proceedings of the IEEE/CVF international conference on computer vision}, pages 1314--1324, 2019.

\bibitem[Howard et~al.(2017)Howard, Zhu, Chen, Kalenichenko, Wang, Weyand, Andreetto, and Adam]{howard2017mobilenets}
Andrew~G Howard, Menglong Zhu, Bo Chen, Dmitry Kalenichenko, Weijun Wang, Tobias Weyand, Marco Andreetto, and Hartwig Adam.
\newblock Mobilenets: Efficient convolutional neural networks for mobile vision applications.
\newblock \emph{arXiv preprint arXiv:1704.04861}, 2017.

\bibitem[Huang et~al.(2022)Huang, Shi, Zhang, Wang, Cheung, Qin, Dai, and Li]{huang2022flowformer}
Zhaoyang Huang, Xiaoyu Shi, Chao Zhang, Qiang Wang, Ka~Chun Cheung, Hongwei Qin, Jifeng Dai, and Hongsheng Li.
\newblock Flowformer: A transformer architecture for optical flow.
\newblock In \emph{Computer Vision--ECCV 2022: 17th European Conference, Tel Aviv, Israel, October 23--27, 2022, Proceedings, Part XVII}, pages 668--685. Springer, 2022.

\bibitem[Hui and Loy(2020)]{hui2020liteflownet3}
Tak-Wai Hui and Chen~Change Loy.
\newblock Liteflownet3: Resolving correspondence ambiguity for more accurate optical flow estimation.
\newblock In \emph{Computer Vision--ECCV 2020: 16th European Conference, Glasgow, UK, August 23--28, 2020, Proceedings, Part XX 16}, pages 169--184. Springer, 2020.

\bibitem[Hui et~al.(2018)Hui, Tang, and Loy]{hui18liteflownet}
Tak-Wai Hui, Xiaoou Tang, and Chen~Change Loy.
\newblock {LiteFlowNet: A Lightweight Convolutional Neural Network for Optical Flow Estimation}.
\newblock In \emph{{Proceedings of IEEE Conference on Computer Vision and Pattern Recognition (CVPR)}}, pages 8981--8989, 2018.

\bibitem[Hur and Roth(2019)]{hur2019iterative}
Junhwa Hur and Stefan Roth.
\newblock Iterative residual refinement for joint optical flow and occlusion estimation.
\newblock In \emph{Proceedings of the IEEE/CVF Conference on Computer Vision and Pattern Recognition}, pages 5754--5763, 2019.

\bibitem[Ignatov et~al.(2018)Ignatov, Timofte, Chou, Wang, Wu, Hartley, and Van~Gool]{ignatov2018ai}
Andrey Ignatov, Radu Timofte, William Chou, Ke Wang, Max Wu, Tim Hartley, and Luc Van~Gool.
\newblock Ai benchmark: Running deep neural networks on android smartphones.
\newblock In \emph{Proceedings of the European Conference on Computer Vision (ECCV) Workshops}, pages 0--0, 2018.

\bibitem[Ilg et~al.(2017)Ilg, Mayer, Saikia, Keuper, Dosovitskiy, and Brox]{ilg2017flownet}
Eddy Ilg, Nikolaus Mayer, Tonmoy Saikia, Margret Keuper, Alexey Dosovitskiy, and Thomas Brox.
\newblock Flownet 2.0: Evolution of optical flow estimation with deep networks.
\newblock In \emph{Proceedings of the IEEE conference on computer vision and pattern recognition}, pages 2462--2470, 2017.

\bibitem[Jaegle et~al.(2021)Jaegle, Borgeaud, Alayrac, Doersch, Ionescu, Ding, Koppula, Zoran, Brock, Shelhamer, et~al.]{jaegle2021perceiver}
Andrew Jaegle, Sebastian Borgeaud, Jean-Baptiste Alayrac, Carl Doersch, Catalin Ionescu, David Ding, Skanda Koppula, Daniel Zoran, Andrew Brock, Evan Shelhamer, et~al.
\newblock Perceiver io: A general architecture for structured inputs \& outputs.
\newblock \emph{arXiv preprint arXiv:2107.14795}, 2021.

\bibitem[Janai et~al.(2018)Janai, Guney, Ranjan, Black, and Geiger]{janai2018unsupervised}
Joel Janai, Fatma Guney, Anurag Ranjan, Michael Black, and Andreas Geiger.
\newblock Unsupervised learning of multi-frame optical flow with occlusions.
\newblock In \emph{Proceedings of the European conference on computer vision (ECCV)}, pages 690--706, 2018.

\bibitem[Jiang et~al.(2021)Jiang, Campbell, Lu, Li, and Hartley]{jiang2021learning}
Shihao Jiang, Dylan Campbell, Yao Lu, Hongdong Li, and Richard Hartley.
\newblock Learning to estimate hidden motions with global motion aggregation.
\newblock In \emph{Proceedings of the IEEE/CVF International Conference on Computer Vision}, pages 9772--9781, 2021.

\bibitem[Kondermann et~al.(2016)Kondermann, Nair, Honauer, Krispin, Andrulis, Brock, Gussefeld, Rahimimoghaddam, Hofmann, Brenner, et~al.]{kondermann2016hci}
Daniel Kondermann, Rahul Nair, Katrin Honauer, Karsten Krispin, Jonas Andrulis, Alexander Brock, Burkhard Gussefeld, Mohsen Rahimimoghaddam, Sabine Hofmann, Claus Brenner, et~al.
\newblock The hci benchmark suite: Stereo and flow ground truth with uncertainties for urban autonomous driving.
\newblock In \emph{Proceedings of the IEEE Conference on Computer Vision and Pattern Recognition Workshops}, pages 19--28, 2016.

\bibitem[Kong and Yang(2020)]{kong2020fdflownet}
Lingtong Kong and Jie Yang.
\newblock Fdflownet: Fast optical flow estimation using a deep lightweight network.
\newblock In \emph{2020 IEEE International Conference on Image Processing (ICIP)}, pages 1501--1505. IEEE, 2020.

\bibitem[Kong et~al.(2021)Kong, Shen, and Yang]{kong2021fastflownet}
Lingtong Kong, Chunhua Shen, and Jie Yang.
\newblock Fastflownet: A lightweight network for fast optical flow estimation.
\newblock In \emph{2021 IEEE International Conference on Robotics and Automation (ICRA)}, pages 10310--10316. IEEE, 2021.

\bibitem[Kroeger et~al.(2016)Kroeger, Timofte, Dai, and Van~Gool]{kroeger2016fast}
Till Kroeger, Radu Timofte, Dengxin Dai, and Luc Van~Gool.
\newblock Fast optical flow using dense inverse search.
\newblock In \emph{Computer Vision--ECCV 2016: 14th European Conference, Amsterdam, The Netherlands, October 11--14, 2016, Proceedings, Part IV 14}, pages 471--488. Springer, 2016.

\bibitem[Lee et~al.(2019{\natexlab{a}})Lee, Chirkov, Ignasheva, Pisarchyk, Shieh, Riccardi, Sarokin, Kulik, and Grundmann]{lee2019device}
Juhyun Lee, Nikolay Chirkov, Ekaterina Ignasheva, Yury Pisarchyk, Mogan Shieh, Fabio Riccardi, Raman Sarokin, Andrei Kulik, and Matthias Grundmann.
\newblock On-device neural net inference with mobile gpus.
\newblock \emph{arXiv preprint arXiv:1907.01989}, 2019{\natexlab{a}}.

\bibitem[Lee and Park(2020)]{lee2020centermask}
Youngwan Lee and Jongyoul Park.
\newblock Centermask: Real-time anchor-free instance segmentation.
\newblock In \emph{Proceedings of the IEEE/CVF conference on computer vision and pattern recognition}, pages 13906--13915, 2020.

\bibitem[Lee et~al.(2019{\natexlab{b}})Lee, Hwang, Lee, Bae, and Park]{lee2019energy}
Youngwan Lee, Joong-won Hwang, Sangrok Lee, Yuseok Bae, and Jongyoul Park.
\newblock An energy and gpu-computation efficient backbone network for real-time object detection.
\newblock In \emph{Proceedings of the IEEE/CVF conference on computer vision and pattern recognition workshops}, pages 0--0, 2019{\natexlab{b}}.

\bibitem[Li et~al.(2022)Li, Yuan, Wen, Hu, Evangelidis, Tulyakov, Wang, and Ren]{li2022efficientformer}
Yanyu Li, Geng Yuan, Yang Wen, Ju Hu, Georgios Evangelidis, Sergey Tulyakov, Yanzhi Wang, and Jian Ren.
\newblock Efficientformer: Vision transformers at mobilenet speed.
\newblock \emph{Advances in Neural Information Processing Systems}, 35:\penalty0 12934--12949, 2022.

\bibitem[Luo et~al.(2023)Luo, Yang, Li, Nie, Lin, Fan, and Liu]{luo2023gaflow}
Ao Luo, Fan Yang, Xin Li, Lang Nie, Chunyu Lin, Haoqiang Fan, and Shuaicheng Liu.
\newblock Gaflow: Incorporating gaussian attention into optical flow.
\newblock In \emph{Proceedings of the IEEE/CVF International Conference on Computer Vision}, pages 9642--9651, 2023.

\bibitem[Mayer et~al.(2016)Mayer, Ilg, Hausser, Fischer, Cremers, Dosovitskiy, and Brox]{mayer2016large}
Nikolaus Mayer, Eddy Ilg, Philip Hausser, Philipp Fischer, Daniel Cremers, Alexey Dosovitskiy, and Thomas Brox.
\newblock A large dataset to train convolutional networks for disparity, optical flow, and scene flow estimation.
\newblock In \emph{Proceedings of the IEEE conference on computer vision and pattern recognition}, pages 4040--4048, 2016.

\bibitem[Mehta and Rastegari(2021)]{mehta2021mobilevit}
Sachin Mehta and Mohammad Rastegari.
\newblock Mobilevit: light-weight, general-purpose, and mobile-friendly vision transformer.
\newblock \emph{arXiv preprint arXiv:2110.02178}, 2021.

\bibitem[Menze and Geiger(2015)]{Menze2015CVPR}
Moritz Menze and Andreas Geiger.
\newblock Object scene flow for autonomous vehicles.
\newblock In \emph{Conference on Computer Vision and Pattern Recognition (CVPR)}, 2015.

\bibitem[Nayman et~al.(2021)Nayman, Aflalo, Noy, and Zelnik]{nayman2021hardcore}
Niv Nayman, Yonathan Aflalo, Asaf Noy, and Lihi Zelnik.
\newblock Hardcore-nas: Hard constrained differentiable neural architecture search.
\newblock In \emph{International Conference on Machine Learning}, pages 7979--7990. PMLR, 2021.

\bibitem[Niklaus and Liu(2018)]{niklaus2018context}
Simon Niklaus and Feng Liu.
\newblock Context-aware synthesis for video frame interpolation.
\newblock In \emph{Proceedings of the IEEE conference on computer vision and pattern recognition}, pages 1701--1710, 2018.

\bibitem[Paszke et~al.(2019)Paszke, Gross, Massa, Lerer, Bradbury, Chanan, Killeen, Lin, Gimelshein, Antiga, Desmaison, Kopf, Yang, DeVito, Raison, Tejani, Chilamkurthy, Steiner, Fang, Bai, and Chintala]{NEURIPS2019_bdbca288}
Adam Paszke, Sam Gross, Francisco Massa, Adam Lerer, James Bradbury, Gregory Chanan, Trevor Killeen, Zeming Lin, Natalia Gimelshein, Luca Antiga, Alban Desmaison, Andreas Kopf, Edward Yang, Zachary DeVito, Martin Raison, Alykhan Tejani, Sasank Chilamkurthy, Benoit Steiner, Lu Fang, Junjie Bai, and Soumith Chintala.
\newblock Pytorch: An imperative style, high-performance deep learning library.
\newblock In \emph{Advances in Neural Information Processing Systems}. Curran Associates, Inc., 2019.

\bibitem[Ranjan and Black(2017)]{ranjan2017optical}
Anurag Ranjan and Michael~J Black.
\newblock Optical flow estimation using a spatial pyramid network.
\newblock In \emph{Proceedings of the IEEE conference on computer vision and pattern recognition}, pages 4161--4170, 2017.

\bibitem[Ren et~al.(2019)Ren, Gallo, Sun, Yang, Sudderth, and Kautz]{8658399}
Zhile Ren, Orazio Gallo, Deqing Sun, Ming-Hsuan Yang, Erik~B. Sudderth, and Jan Kautz.
\newblock A fusion approach for multi-frame optical flow estimation.
\newblock In \emph{2019 IEEE Winter Conference on Applications of Computer Vision (WACV)}, pages 2077--2086, 2019.

\bibitem[Richter et~al.(2017)Richter, Hayder, and Koltun]{Richter_2017}
Stephan~R. Richter, Zeeshan Hayder, and Vladlen Koltun.
\newblock Playing for benchmarks.
\newblock In \emph{{IEEE} International Conference on Computer Vision, {ICCV} 2017, Venice, Italy, October 22-29, 2017}, pages 2232--2241, 2017.

\bibitem[Sandler et~al.(2018)Sandler, Howard, Zhu, Zhmoginov, and Chen]{sandler2018mobilenetv2}
Mark Sandler, Andrew Howard, Menglong Zhu, Andrey Zhmoginov, and Liang-Chieh Chen.
\newblock Mobilenetv2: Inverted residuals and linear bottlenecks.
\newblock In \emph{Proceedings of the IEEE conference on computer vision and pattern recognition}, pages 4510--4520, 2018.

\bibitem[Shi et~al.(2023{\natexlab{a}})Shi, Huang, Bian, Li, Zhang, Cheung, See, Qin, Dai, and Li]{shi2023videoflow}
Xiaoyu Shi, Zhaoyang Huang, Weikang Bian, Dasong Li, Manyuan Zhang, Ka~Chun Cheung, Simon See, Hongwei Qin, Jifeng Dai, and Hongsheng Li.
\newblock Videoflow: Exploiting temporal cues for multi-frame optical flow estimation.
\newblock \emph{arXiv preprint arXiv:2303.08340}, 2023{\natexlab{a}}.

\bibitem[Shi et~al.(2023{\natexlab{b}})Shi, Huang, Li, Zhang, Cheung, See, Qin, Dai, and Li]{shi2023flowformer++}
Xiaoyu Shi, Zhaoyang Huang, Dasong Li, Manyuan Zhang, Ka~Chun Cheung, Simon See, Hongwei Qin, Jifeng Dai, and Hongsheng Li.
\newblock Flowformer++: Masked cost volume autoencoding for pretraining optical flow estimation.
\newblock \emph{arXiv preprint arXiv:2303.01237}, 2023{\natexlab{b}}.

\bibitem[Shuang et~al.(2020)Shuang, Huang, Sun, Cai, and Guo]{shuang2020fine}
Kai Shuang, Yuheng Huang, Yue Sun, Zhun Cai, and Hao Guo.
\newblock Fine-grained motion representation for template-free visual tracking.
\newblock In \emph{Proceedings of the IEEE/CVF Winter Conference on Applications of Computer Vision}, pages 671--680, 2020.

\bibitem[Simonyan and Zisserman(2014)]{simonyan2014recognition}
Karen Simonyan and Andrew Zisserman.
\newblock Two-stream convolutional networks for action recognition in videos.
\newblock In \emph{Proceedings of the 27th International Conference on Neural Information Processing Systems - Volume 1}, page 568–576, Cambridge, MA, USA, 2014. MIT Press.

\bibitem[Sun et~al.(2018)Sun, Yang, Liu, and Kautz]{sun2018pwc}
Deqing Sun, Xiaodong Yang, Ming-Yu Liu, and Jan Kautz.
\newblock Pwc-net: Cnns for optical flow using pyramid, warping, and cost volume.
\newblock In \emph{Proceedings of the IEEE conference on computer vision and pattern recognition}, pages 8934--8943, 2018.

\bibitem[Sun et~al.(2019)Sun, Yang, Liu, and Kautz]{sun2019models}
Deqing Sun, Xiaodong Yang, Ming-Yu Liu, and Jan Kautz.
\newblock Models matter, so does training: An empirical study of cnns for optical flow estimation.
\newblock \emph{IEEE transactions on pattern analysis and machine intelligence}, 42\penalty0 (6):\penalty0 1408--1423, 2019.

\bibitem[Sun et~al.(2021)Sun, Vlasic, Herrmann, Jampani, Krainin, Chang, Zabih, Freeman, and Liu]{sun2021autoflow}
Deqing Sun, Daniel Vlasic, Charles Herrmann, Varun Jampani, Michael Krainin, Huiwen Chang, Ramin Zabih, William~T Freeman, and Ce Liu.
\newblock Autoflow: Learning a better training set for optical flow.
\newblock In \emph{Proceedings of the IEEE/CVF Conference on Computer Vision and Pattern Recognition}, pages 10093--10102, 2021.

\bibitem[Sun et~al.(2022)Sun, Herrmann, Reda, Rubinstein, Fleet, and Freeman]{sun2022disentangling}
Deqing Sun, Charles Herrmann, Fitsum Reda, Michael Rubinstein, David~J Fleet, and William~T Freeman.
\newblock Disentangling architecture and training for optical flow.
\newblock In \emph{Computer Vision--ECCV 2022: 17th European Conference, Tel Aviv, Israel, October 23--27, 2022, Proceedings, Part XXII}, pages 165--182. Springer, 2022.

\bibitem[Tan and Le(2019{\natexlab{a}})]{tan2019efficientnet}
Mingxing Tan and Quoc Le.
\newblock Efficientnet: Rethinking model scaling for convolutional neural networks.
\newblock In \emph{International conference on machine learning}, pages 6105--6114. PMLR, 2019{\natexlab{a}}.

\bibitem[Tan and Le(2021)]{tan2021efficientnetv2}
Mingxing Tan and Quoc Le.
\newblock Efficientnetv2: Smaller models and faster training.
\newblock In \emph{International conference on machine learning}, pages 10096--10106. PMLR, 2021.

\bibitem[Tan and Le(2019{\natexlab{b}})]{tan2019mixconv}
Mingxing Tan and Quoc~V Le.
\newblock Mixconv: Mixed depthwise convolutional kernels.
\newblock \emph{arXiv preprint arXiv:1907.09595}, 2019{\natexlab{b}}.

\bibitem[Tan et~al.(2019)Tan, Chen, Pang, Vasudevan, Sandler, Howard, and Le]{tan2019mnasnet}
Mingxing Tan, Bo Chen, Ruoming Pang, Vijay Vasudevan, Mark Sandler, Andrew Howard, and Quoc~V Le.
\newblock Mnasnet: Platform-aware neural architecture search for mobile.
\newblock In \emph{Proceedings of the IEEE/CVF conference on computer vision and pattern recognition}, pages 2820--2828, 2019.

\bibitem[Teed and Deng(2020)]{teed2020raft}
Zachary Teed and Jia Deng.
\newblock Raft: Recurrent all-pairs field transforms for optical flow.
\newblock In \emph{Computer Vision--ECCV 2020: 16th European Conference, Glasgow, UK, August 23--28, 2020, Proceedings, Part II 16}, pages 402--419. Springer, 2020.

\bibitem[Wadekar and Chaurasia(2022)]{wadekar2022mobilevitv3}
Shakti~N. Wadekar and Abhishek Chaurasia.
\newblock Mobilevitv3: Mobile-friendly vision transformer with simple and effective fusion of local, global and input features.
\newblock 2022.

\bibitem[Wang et~al.(2018)Wang, Ji, Nguyen, and Xie]{wang2018correlation}
Chen Wang, Tete Ji, Thien-Minh Nguyen, and Lihua Xie.
\newblock Correlation flow: robust optical flow using kernel cross-correlators.
\newblock In \emph{2018 IEEE International Conference on Robotics and Automation (ICRA)}, pages 836--841. IEEE, 2018.

\bibitem[Wu et~al.(2019)Wu, Dai, Zhang, Wang, Sun, Wu, Tian, Vajda, Jia, and Keutzer]{wu2019fbnet}
Bichen Wu, Xiaoliang Dai, Peizhao Zhang, Yanghan Wang, Fei Sun, Yiming Wu, Yuandong Tian, Peter Vajda, Yangqing Jia, and Kurt Keutzer.
\newblock Fbnet: Hardware-aware efficient convnet design via differentiable neural architecture search.
\newblock In \emph{Proceedings of the IEEE/CVF Conference on Computer Vision and Pattern Recognition}, pages 10734--10742, 2019.

\bibitem[Xu et~al.(2021)Xu, Yang, Cai, Zhang, and Tong]{xu2021high}
Haofei Xu, Jiaolong Yang, Jianfei Cai, Juyong Zhang, and Xin Tong.
\newblock High-resolution optical flow from 1d attention and correlation.
\newblock In \emph{Proceedings of the IEEE/CVF International Conference on Computer Vision}, pages 10498--10507, 2021.

\bibitem[Xu et~al.(2022)Xu, Zhang, Cai, Rezatofighi, and Tao]{xu2022gmflow}
Haofei Xu, Jing Zhang, Jianfei Cai, Hamid Rezatofighi, and Dacheng Tao.
\newblock Gmflow: Learning optical flow via global matching.
\newblock In \emph{Proceedings of the IEEE/CVF conference on computer vision and pattern recognition}, pages 8121--8130, 2022.

\bibitem[Xu et~al.(2017)Xu, Ranftl, and Koltun]{xu2017accurate}
Jia Xu, Ren{\'e} Ranftl, and Vladlen Koltun.
\newblock Accurate optical flow via direct cost volume processing.
\newblock In \emph{Proceedings of the IEEE Conference on Computer Vision and Pattern Recognition}, pages 1289--1297, 2017.

\bibitem[Xue et~al.(2019)Xue, Chen, Wu, Wei, and Freeman]{xue2019video}
Tianfan Xue, Baian Chen, Jiajun Wu, Donglai Wei, and William~T Freeman.
\newblock Video enhancement with task-oriented flow.
\newblock \emph{International Journal of Computer Vision}, 127:\penalty0 1106--1125, 2019.

\bibitem[Yang and Ramanan(2019)]{yang2019volumetric}
Gengshan Yang and Deva Ramanan.
\newblock Volumetric correspondence networks for optical flow.
\newblock \emph{Advances in neural information processing systems}, 32, 2019.

\bibitem[Yang et~al.(2021)Yang, Kong, and Yang]{yang2021unsupervised}
Xiaohang Yang, Lingtong Kong, and Jie Yang.
\newblock Unsupervised motion representation enhanced network for action recognition.
\newblock In \emph{ICASSP 2021-2021 IEEE International Conference on Acoustics, Speech and Signal Processing (ICASSP)}, pages 2445--2449. IEEE, 2021.

\bibitem[Yu and Ramamoorthi(2020)]{yu2020learning}
Jiyang Yu and Ravi Ramamoorthi.
\newblock Learning video stabilization using optical flow.
\newblock In \emph{Proceedings of the IEEE/CVF Conference on Computer Vision and Pattern Recognition}, pages 8159--8167, 2020.

\bibitem[Zach et~al.(2007)Zach, Pock, and Bischof]{zach2007duality}
Christopher Zach, Thomas Pock, and Horst Bischof.
\newblock A duality based approach for realtime tv-l 1 optical flow.
\newblock In \emph{Pattern Recognition: 29th DAGM Symposium, Heidelberg, Germany, September 12-14, 2007. Proceedings 29}, pages 214--223. Springer, 2007.

\bibitem[Zhang et~al.(2021)Zhang, Woodford, Prisacariu, and Torr]{zhang2021separable}
Feihu Zhang, Oliver~J Woodford, Victor~Adrian Prisacariu, and Philip~HS Torr.
\newblock Separable flow: Learning motion cost volumes for optical flow estimation.
\newblock In \emph{Proceedings of the IEEE/CVF International Conference on Computer Vision}, pages 10807--10817, 2021.

\bibitem[Zhao et~al.(2022)Zhao, Zhao, Zhang, Zhou, and Metaxas]{zhao2022global}
Shiyu Zhao, Long Zhao, Zhixing Zhang, Enyu Zhou, and Dimitris Metaxas.
\newblock Global matching with overlapping attention for optical flow estimation.
\newblock In \emph{Proceedings of the IEEE/CVF Conference on Computer Vision and Pattern Recognition}, pages 17592--17601, 2022.

\end{thebibliography}
}


\end{document}